\documentclass[10pt,twocolumn,letterpaper]{article}

\usepackage{wacv}              

\usepackage{graphicx}
\usepackage{amsmath}
\usepackage{amssymb}
\usepackage{booktabs}
\usepackage[accsupp]{axessibility}

\usepackage[pagebackref,breaklinks,colorlinks]{hyperref}

\usepackage[capitalize]{cleveref}
\crefname{section}{Sec.}{Secs.}
\Crefname{section}{Section}{Sections}
\Crefname{table}{Table}{Tables}
\crefname{table}{Tab.}{Tabs.}

\usepackage{comment}

\def\wacvPaperID{518} 
\def\confName{WACV}
\def\confYear{2025}

\begin{document}

\title{uLayout: Unified Room Layout Estimation for Perspective and \\ Panoramic Images}

\author{
Jonathan Lee$^{1}$\quad
Bolivar Solarte$^{1}$\quad
Chin-Hsuan Wu$^{1}$\quad
Jin-Cheng Jhang$^{1}$\quad
Fu-En Wang$^{1}$\quad
\\
Yi-Hsuan Tsai$^{2}$\quad
Min Sun$^{1}$\quad
\\
$^{1}$National Tsing Hua University, Taiwan\quad
$^{2}$Atmanity Inc.\quad
\\
{\tt\small \{jonathanlee19896, enrique.solarte.pardo, wasidennis\}@gmail.com}
\\
{\tt\small \{chinhsuanwu, frank890725, fulton84717\}@gapp.nthu.edu.tw}\quad
{\tt\small sunmin@ee.nthu.edu.tw}
}

\definecolor{red}{rgb}{0.9,0.1,0}
\definecolor{gray}{rgb}{0.8,0.8,0.8}
\definecolor{blue}{rgb}{0.4,0.4,0.9}
\definecolor{green}{rgb}{0, 0.4, 0}
\definecolor{orange}{rgb}{1, 0.5, 0}
\definecolor{slateblue}{rgb}{0.7,0.35,0.9}
\definecolor{mahogany}{rgb}{0.75, 0.25, 0.0}
\definecolor{purple}{rgb}{0.6, 0, 0.6}
\definecolor{goldenrod}{rgb}{0.85, 0.65, 0.13}
\newbool{revising}
\setbool{revising}{false}
\ifbool{revising}
{
    \newcommand{\iref}[1]{{#1}}
    \newcommand{\todo}[1]{{\color{red}#1}}
    \newcommand{\kike}[1]{\textcolor{blue}{#1}}
     \newcommand{\kikecomment}[1]
     {\textcolor{green}{[kike: #1]}}
    \newcommand{\justin}[1]{\textcolor{green}{#1}}
    \newcommand{\justincomment}[1]{\textcolor{green}{[justin: #1]}}
    \newcommand{\frank}[1]{\textcolor{slateblue}{#1}}
    \newcommand{\jonathan}[1]{\textcolor{mahogany}{#1}}
    \newcommand{\fuen}[1]{\textcolor{purple}{#1}}
    \newcommand{\dennis}[1]{\textcolor{orange}{#1}}
    \newcommand{\minsun}[1]{\textcolor{magenta}{#1}}
    \newcommand{\idea}[1]{\textcolor{red}{[Idea]:#1}}
} {
    \newcommand{\iref}[1]{\textcolor{blue}{}}
    \newcommand{\kike}[1]{#1}
    \newcommand{\justin}[1]{{#1}}
    \newcommand{\frank}[1]{{#1}}
    \newcommand{\jonathan}[1]{{#1}}
    \newcommand{\fuen}[1]{{#1}}
    \newcommand{\dennis}[1]{{#1}}
    \newcommand{\minsun}[1]{{#1}}
    \newcommand{\todo}[1]{{}}
    \newcommand{\TODO}[1]{{}}
    \newcommand{\idea}[1]{{}} 
}
\newcommand{\dummyfig}[1]{
  \centering
  \fbox{
    \begin{minipage}[c][0.25\textheight][c]{0.5\textwidth}
      \centering{#1}
    \end{minipage}
  }
}

\maketitle
\begin{abstract}
\kike{We present uLayout, a unified model for estimating room layout geometries from both perspective and panoramic images, whereas traditional solutions require different model designs for each image type.}
\kike{The key idea of our solution is to unify both domains into the equirectangular projection, particularly, allocating perspective images into the most suitable latitude coordinate to effectively exploit both domains seamlessly.}
\kike{To address the Field-of-View (FoV) difference between the input domains, we design uLayout with a shared feature extractor with an extra 1D-Convolution layer to condition each domain input differently. This conditioning allows us to efficiently formulate a column-wise feature regression problem regardless of the FoV input.}
\kike{This simple yet effective approach achieves competitive performance with current state-of-the-art solutions and shows for the first time a single end-to-end model for both domains.
Extensive experiments in the real-world datasets, LSUN, Matterport3D, PanoContext, and Stanford 2D-3D evidence the contribution of our approach. Code is available at
\href{https://github.com/JonathanLee112/uLayout}{ https://github.com/JonathanLee112/uLayout}.}

\end{abstract}
\begin{figure*}[t!]
\centering
  \includegraphics[width=\linewidth]
    {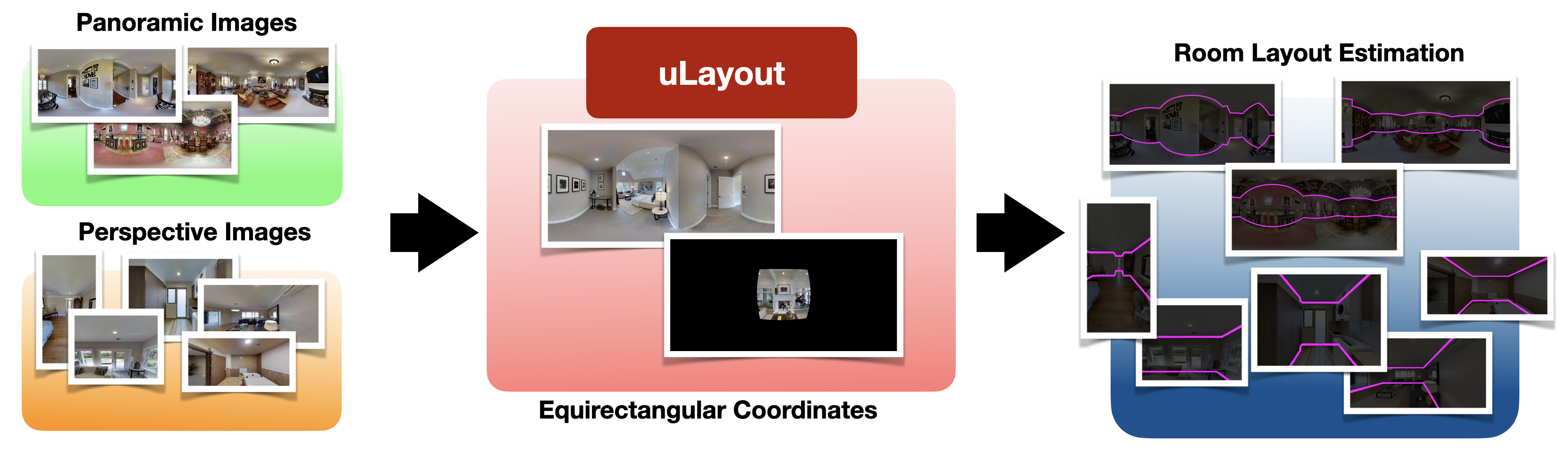}
    \caption{\textbf{Method Overview}. uLayout is a unified layout estimation model jointly trained with panoramic and perspective images for predicting ceiling and floor boundaries.
    uLayout is designed to efficiently handle images with different Field-of-Views (FoV) (See comparison in equirectangular coordinate in the middle) in both training and inference. uLayout achieves highly competitive performance on LSUN~\cite{lsun_dataset} and Matterport3D~\cite{matterport3d} datasets.}
    \label{fig_teaser}
\end{figure*}

\section{Introduction}
\kike{The most simple geometry representation of an indoor scene is its room layout geometry, which plays a crucial role in many downstream tasks such as scene understanding, robot navigation, and 3D-reconstruction~\cite{nie2022pose2room, 360_dfpe, chen20224dcontrast}.}
\kike{For this task, the most practical and stable solutions have been those based on perspective and panorama images~\cite{lsun_room_2018, fusing2023, horizonnet, jiang2022lgt}.}
\kike{However, these approaches present significant differences in design, resulting in a limitation to exploit both data domains with a unique model.}

\kike{Solutions for room layout estimation using perspective images typically define the problem as a semantic segmentation task, aiming to classify image segments into walls, floor, and ceiling~\cite{hedau2009, lsun_room_2018, fusing2023}. In contrast, panorama-based solutions leverage the equirectangular image projection to define the layout estimation as column-wise feature regression 
thus avoiding the semantic classification problem~\cite{horizonnet, jiang2022lgt, yang2019dula, wang2021led2, tsai2024no, solarte2024self}.}
\kike{To the best of our knowledge, a layout estimation model that handles perspective and panorama image domains has not been proposed.}

\kike{In this paper, we propose uLayout, a unified solution for room layout estimation that handles both perspective and panoramic images.}
\kike{Our solution projects both input data into equirectangular coordinates, to define a column-wise feature regression for both domains. Particularly, we project perspective images into a specific latitude coordinate by assuming that a pitch angle orientation is given.}

\kike{Additionally, to efficiently address the FoV differences between the domain inputs,
we use a shared feature extractor for both domains, with each domain being conditioned differently using extra 1D convolution layers.}
\kike{These conditioning layers allow us to adapt the difference in the sequence length of column-wise features due to differences in the FoV inputs. This simple yet effective approach enables us to learn from both domains effectively and efficiently.}

\kike{We corroborate the benefits of uLayout in extensive experiments with different setting evaluations on publicly available real-world datasets~\cite{lsun_dataset, mp3d_dataset, zhang2014panocontext, armeni2017st2d3d}. We compared with the current state-of-the-art solutions LSUN-ROOM~\cite{lsun_room_2018} and FUSING~\cite{fusing2023} for perspective images. We also compare with the current state-of-the-art solutions LGT-Net~\cite{jiang2022lgt} and DOP-Net~\cite{shen2023dopnet} for panoramic images. Our results show for the first time competitive performance in both domains simultaneously. The details of our contribution can be listed as follows:}
\kike{
\begin{itemize}
    \item We present uLayout, a novel formulation for room layout estimation that handles both perspective and panoramic images as input.
    \item We efficiently use the equirectangular projection for both input domains, particularly addressing the difference in the FoV and the allocation of perspective images in the equirectangular coordinates. 
    \item Through extensive experiments in real-world datasets, we show for the first time state-of-art performance in both domains simultaneously.
\end{itemize}
}
\section{Related Work}
\label{sec:related_work}

\subsection{Perspective Layout Estimation} 
\jonathan{
The one of the early perspective layout estimation work was introduced by~\cite{hedau2009}. 
After deep learning approaches appeared, methods utilizing Convolutional Neural Networks (CNN) and Fully Convolutional Networks (FCN) have been proposed for layout estimation.~\cite{mallya2015} was the first to introduce FCN as a segmentation network to layout estimation by predicting informative edge maps.~\cite{dasgupta2016} proposed the promising two-stage network for segmenting the planes and walls by deep learning method and optimizing the output of the vanishing points. To further improve the performance in LSUN dataset~\cite{lsun_dataset}, 
LSUN-ROOM~\cite{lsun_room_2018} noted that room types with fewer images in LSUN training datasets result in higher pixel errors, particularly when surfaces are few. To address this issue, they propose layout structure degeneration, which transforms room types with more surfaces into those with fewer surfaces. After obtaining sufficient training data in each type, the pixel error can be significantly reduced, rendering this model one of the best 2D layout estimation models. For this reason, we have selected LSUN-ROOM~\cite{lsun_room_2018} as one of our baseline models.}

\jonathan{
In addition to focusing on 2D layout estimation, some researchers have begun addressing the task of estimating 3D layouts.~\cite{lee2009} proposed the ``Indoor World'' model, based on the Manhattan world assumption and symmetric floor and ceiling principles. This 2D model effectively represents scene layouts and can be converted into a 3D representation using geometric analysis of edge configurations.~\cite{zhang2020} extract the pixel-level surface parameters of the dominant planes, e.g., ceiling, floor, and walls, from the depth map and utilize the deep neural network to learn the surface parameters and estimate layout by intersecting the depth maps generated from surface parameters. 
To further improve the performance,
FUSING~\cite{fusing2023} employs the depth learning module to generate a depth map, augmenting it with virtual planar obstacles to mask specific regions. Incorporating a depth learning module not only addresses the absence of 3D data in 2D images but also effectively captures the relationship between indoor objects and spatial structures. Following this, feature maps are extracted using a structural feature extraction module and then fused with those obtained from the encoder of the depth learning module. This integrated approach, leveraging both color and depth information, enables the model to predict surface parameters at the pixel level with higher quality compared to~\cite{zhang2020}, establishing it as one of the top-performing 3D layout estimation models.}

\jonathan{
Moreover, FUSING is primarily trained on both Matterport3D dataset~\cite{matterport3d} and the LSUN dataset~\cite{lsun_dataset}, aligning with our dataset configuration. The only distinction is that our Matterport3D data is panoramic, while theirs is perspective.  
Considering these factors, we select FUSING as one of the baseline models for our study. In summary, our baselines include LSUN-ROOM and FUSING, representing 2D and 3D-based perspective state-of-the-art methods, respectively.}

\subsection{Panoramic Layout Estimation}
\jonathan{
In the realm of panoramic layout estimation, previous methodologies have adhered to the Manhattan World assumption~\cite{coughlan1999manhattan}. For example, LayoutNet~\cite{zou2018layoutnet} directly predicts corner and boundary probability maps from panoramas. Dula-Net~\cite{yang2019dula} predicts semantic masks for 2D floor planes using equirectangular and perspective views of ceilings. EquiConvs~\cite{fernandez2020corners} employed equirectangular convolutions to generate corner and edge probability maps. HorizonNet~\cite{horizonnet} and HoHoNet~\cite{sun2021hohonet} simplify layout estimation by utilizing 1D representations and employing Bi-LSTM and multi-head self-attention mechanisms~\cite{vaswani2017attention} to establish long-range dependencies. LE$\mathrm{D}^{2}$-Net~\cite{wang2021led2} reframes layout estimation by predicting the depth of walls horizontally. AtlantaNet~\cite{pintore2020atlantanet} predicts room layouts by amalgamating projections of floor and ceiling planes. DMH-Net~\cite{zhao20223d} transforms panoramas into cubemaps and predicts the position of intersection lines with a learnable Hough Transform Block. LGT-Net~\cite{jiang2022lgt} utilizes self-attention transformers to learn geometric relationships and capture long-range dependencies. DOP-Net~\cite{shen2023dopnet} disentangles 1D features by segmenting them into orthogonal plane representations and refines the features using Graph Convolutional Network(GCN) and transformer mechanisms.}

\section{Method}
In this section, we present the details of our uLayout approach for unifying room layout estimation for perspective and panorama images. In~\cref{sec:align_pano_pp_layout}, we present details of our preprocessing stage that allows us to align both perspective and panoramic images into equirectangular coordinates for a unified image projection.
In~\cref{sec:vertical_shift}, we introduce an image allocation process, which aims to allocate perspective images onto an equirectangular projection by adjusting the vertical position based on pitch angle orientation.
In~\cref{sec:ulayout}, we detail our uLayout model architecture and the loss function employed to constrain both image domains successfully.
An overview of our formulation is depicted in~\cref{fig_model_structure}.

\subsection{Image Preprocessing}
\label{sec:align_pano_pp_layout}


\kike{In the panorama realm, a common preprocessing technique is the vertical image alignment which ensures that all vertical structures are aligned to the Y-axis in the equirectangular coordinates, ensuring that the layout estimation problem can be defined as a column-wise feature regression~\cite{horizonnet}. In this work, we vertically align both perspective and panorama images following the procedure described in~\cite{zhang2014panocontext}.}

\kike{Additionally, to address the difference between both domain inputs due to different camera models, we project both perspective and panorama images into the equirectangular projection. For panorama images, we follow the standard procedure presented in~\cite{horizonnet, sun2021hohonet, wang2021led2, jiang2022lgt, shen2023dopnet}. To project perspective images, we assume that each image view represents the front face of a cubemap projection~\cite{heidrich1998cubemap}. This assumption allows us to directly map the view into equirectangular coordinates.}

\subsection{Allocation Process for Perspective Images}
\label{sec:vertical_shift}
\begin{figure}[ht]
\centering
  \includegraphics[ width=\linewidth]
    {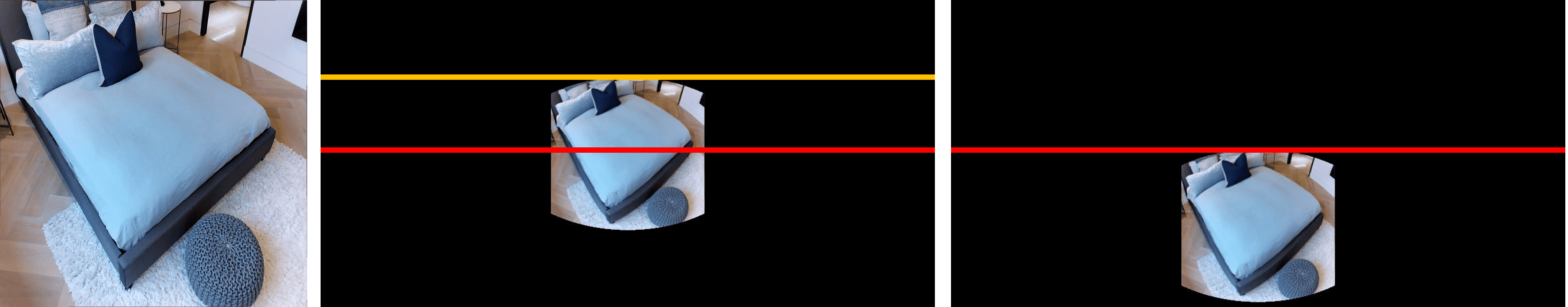}
    \begin{picture}(0, 0)
    \put(-100,3){(a)}
    \put(-28,3){(b)}
    \put(68,3){(c)}
    \end{picture}
    \vspace{-3mm}
    \caption{\textbf{Vertical shift.} Starting with a perspective image (a), we project the image into panoramic coordinates (b) where the yellow line represents the current horizon of the perspective image. We then adjust the pitch orientation by shifting the image downwards to align the yellow line (horizon in perspective) with the red line (desired horizon in the panorama) as shown in (c).
    }
    \label{fig_vertical_shift}
\end{figure}
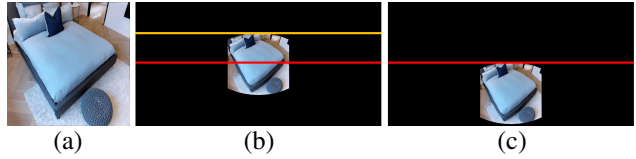
\kike{Despite unifying perspective and panorama images into a common equirectangular projection,
a substantial gap remains between both domains.}
\kike{For instance, panoramic images can visualize the entire room context capturing the ceiling and floor geometries regardless of the camera position.}
\kike{In contrast, perspective images, constrained by a narrower FoV, capture only a portion of the room, which may exclude the ceiling or floor geometries, leading to an ill-posed layout problem.}

\kike{To address this issue, we use the pitch angle orientation of a perspective image, as a given extrinsic parameter, allowing us to align the horizon lines of both the equirectangular and perspective views. 
\jonathan{This alignment process, referred to as ``Vertical Shift''} 
allocates the perspective image into a specific latitude coordinate, asserting that negative and positive pitch orientations place images below and above the equatorial line in the equirectangular view. The examples are illustrated in~\cref{fig_vertical_shift}.}



\kike{Note that for practical applications, a ubiquitous inertial measurement unit (IMU) can be used to acquire pitch angle orientations for perspective images, making this image allocation process both practical and feasible.}
\begin{figure*}[t!]
\centering
  \includegraphics[ width=\linewidth]
    {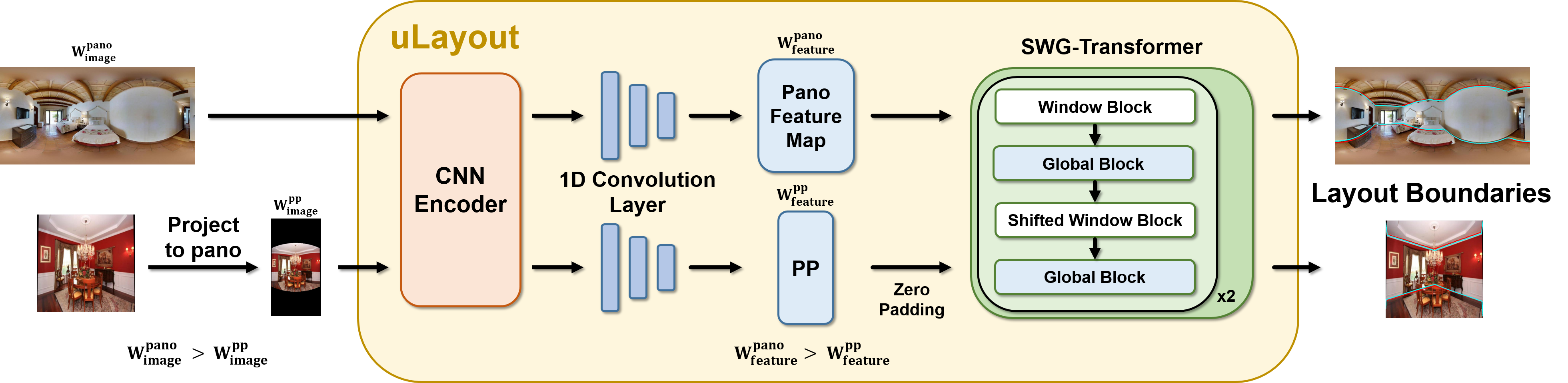}
    \caption{\textbf{Architecture of uLayout.} Firstly, a perspective image is mapped into the same equirectangular coordinate as a panoramic image and applied the reduction method proposed in~\cref{sec:effi_feature_extract}. Therefore, the image width of the perspective image is much smaller than the width of the panoramic image ($w_{image}^{pano} >w_{image}^{pp}$). As a result, the feature maps after CNN and 1D convolution layers have different sizes in the horizontal dimension ($w_{feature}^{pano} >w_{feature}^{pp}$). Hence, the FLOPs and memory usage of perspective images for CNN and 1D convolution computation are much smaller than panoramic images. Additionally, a dual-branch design allows for seamless feature extraction regardless of the different Fields-of-View (FoV) in perspective and panoramic domains.
    Finally, We utilize the SWG-Transformer as our framework and estimate the boundary for both domains, as described in~\cref{sec:swg-transformer} and~\cref{sec:prediction_and_loss}.}
    \label{fig_model_structure}
\end{figure*}
\subsection{uLayout Model}
\label{sec:ulayout}

\jonathan{
For our uLayout model, we introduce details of our proposed feature extractor that effectively exploits both domains with a shared model architecture in~\cref{sec:effi_feature_extract} and describe the architecture we used in~\cref{sec:swg-transformer}, as well as our model prediction and loss settings for panoramic and perspective images in~\cref{sec:prediction_and_loss}.
}
\vspace{-3mm}
\subsubsection{Efficient Feature Extractor}
\label{sec:effi_feature_extract}


\jonathan{
After projecting, aligning, and applying vertical shift to the perspective image, we successfully unify both domains. Due to the limited FoV of the perspective image, projecting perspective coordinates into equirectangular ones produces large non-informative columns, depicted as black regions in~\cref{fig_vertical_shift}-(b), which are irrelevant for our purposes. These regions cause the model to learn non-essential information, wasting time during both training and inference.}

\jonathan{
To address the issue, we propose a new mechanism to minimize the non-informative sections and efficiently extract meaningful information. Specifically, we crop out all non-informative columns in the perspective data similar to the image annotated 
with ($w_{image}^{pp}$ in~\cref{fig_model_structure}), while keeping the whole vertical FoV. This allows us to retain the vertical information while mimicking the cropped panorama data, albeit without the top and bottom information. By employing this reduction method, we can decrease the perspective input image width by 75\%, reducing it from 1024 to 256 columns.
}

\jonathan{
To further optimize data processing, we introduce a dual-branch architecture that incorporates a shared ResNet-50~\cite{Resnet} along with separate convolution layers for each branch. The shared feature extractor allows the model to seamlessly learn features from both domains, while the separate convolutions adapt these features to handle the different sequence lengths that result from cropping perspective images.
}
\jonathan{
For the separate 1D convolution layers, we adopt the design from~\cite{horizonnet}, employing three convolution layers with kernel sizes and strides of 4$\times$1, 2$\times$1, 2$\times$1, applied independently to each branch. These layers capture both low-level and high-level features, compressing the height of the 2D feature maps, which are generated by ResNet-50~\cite{Resnet} at four different scales. Additionally, for panoramic feature maps, we maintain circular consistency by appending the last column before and the first column after each feature map, as proposed in~\cite{horizonnet}. Finally, the feature maps are upsampled to a width of 256 for panoramic maps and 64 for perspective maps and merged into one 2D feature map, each forming a distinct 2D feature map for their respective domains.}

\jonathan{
By applying the reduction and dual-branch architecture, we can adapt extracted features for both domains regardless of the FoV and save approximately 75\% memory usage, and Floating-Point Operations (FLOPs) during the feature extraction process shown in~\cref{sec:ablation}.
The final concatenated feature map sizes are 1$\times$1024$\times$256 for panoramic images and 1$\times$1024$\times$64 for perspective images. Before feeding the data into the transformer, we zero-pad the perspective feature map to match the dimensions of the panoramic feature map, and then concatenate all the feature maps from both domains.
}
\vspace{-3mm}
\subsubsection{SWG-Transformer}
\label{sec:swg-transformer}

\jonathan{
We employ the SWG-Transformer framework as proposed by LGT-Net~\cite{jiang2022lgt}. This framework consists of four sequential blocks: Window Block, Global Block, Shifted Window Block, and Global Block. By default, this sequence is repeated twice, resulting in a total of eight blocks. The Window Block enhances local geometric relationships by dividing the feature maps into several regions. The Shifted Window Block serves a similar purpose but shifts the feature maps before partitioning to capture different regions' local geometric relationships. The Global Block, on the other hand, captures global geometric relationships without any partitioning. By applying this framework, we can seamlessly extract both global and local contextual information and relationships.
}
\vspace{-3mm}
\subsubsection{Prediction and Loss}
\label{sec:prediction_and_loss}

\jonathan{
In the LGT-Net~\cite{jiang2022lgt} setting, the horizon depth from the floor boundary and room height are selected as model predictions. However, due to the limitations of the FoV in perspective images, which often capture only a portion of the room and frequently exclude significant contextual elements like the ceiling or floor, obtaining accurate room height information based on a single ceiling or floor boundary is challenging. 
Therefore, we directly choose the boundaries in the image domain as our model output, avoiding the difficulties associated with calculating room height and dealing with the unsymmetric boundary problem in perspective images.
After extracting both global and local contextual information and relationships with the SWG-Transformer, we apply a linear layer to refine these features, generating the final 1D boundary predictions for the ceiling and floor.
}

\jonathan{
In the panorama loss setting, we apply L1 loss to the boundaries $L_{b}$ in the image domain and the horizon depths $L_{d}$ after projecting the boundaries into a bird's-eye view plane. Additionally, we utilize the normal loss $L_{n}$ and gradient loss $L_{g}$ functions as proposed in LGT-Net~\cite{jiang2022lgt}. The panoramic domain loss function $L_{pano}$ is: 
\begin{equation}
    L_{pano} = \lambda L_{b} +
    \mu L_{d} + \gamma (L_{n} + L_{g})~,
\end{equation}
where $\lambda$, $\mu$, and $\gamma$ are hyper-parameters. In the perspective loss setting, we only apply L1 loss to the boundaries $L_{b}$ in the image domain. The perspective domain loss function $L_{pp}$ is:
\begin{equation}
    L_{pp} = \delta L_{b}~,
\end{equation}
where $\delta$ is a hyper-parameter. The total loss $L_{total}$ applied in the whole model is:
\begin{equation}
    L_{total} = L_{pano} + L_{pp}~.
\end{equation}
}
\vspace{-9mm}


\section{Experiment}

\jonathan{
In this section, we will delve into our experimental setup. In~\cref{sec:exp_setting}, we outline the specifics of our experiment configuration. In~\cref{sec:datasets}, we introduce the datasets utilized for joint training of our uLayout model. In~\cref{sec:evaluation}, we explain the evaluation metrics employed for both panoramic and perspective analyses. In~\cref{sec:experiment}, we present our model's performance, examining its generalization with various dataset combinations. Additionally, we conduct a comparative analysis against two state-of-the-art panoramic models LGT-Net~\cite{jiang2022lgt}, DOP-Net~\cite{shen2023dopnet} and two baseline models for perspective layout estimation: LSUN-ROOM~\cite{lsun_room_2018}, trained on LSUN dataset~\cite{lsun_dataset}, and FUSING~\cite{fusing2023}, trained on perspective data in Matterport3D~\cite{matterport3d} and LSUN dataset~\cite{lsun_dataset}.
}

\jonathan{
In~\cref{sec:ablation}, we will present the results of joint training on both panoramic and perspective datasets and compare them to the results of training on each dataset individually. Additionally, We will explore the impact of our two model designs: vertical shift (\cref{sec:vertical_shift}) and efficient feature extraction (\cref{sec:effi_feature_extract}) on the overall model performance and efficacy.
}


\subsection{Experiment Setting}
\label{sec:exp_setting}

\jonathan{
We utilize PyTorch\cite{pytorch} for implementation. We employ the Adam optimizer with a setting of $\beta_1$ = 0.9 and $\beta_2$ = 0.999 to train the network on a single RTX 4090 GPU for 1000 epochs using a batch size of 16 and a learning rate of 0.0001. For data augmentation in panorama, we adopt the same approaches as used in~\cite{horizonnet}. These include standard left-right flipping, panoramic horizontal rotation, luminance change, and pano-stretch during training. In perspective, we incorporate luminance change and left-right flipping for data augmentation as well. Additionally, we set hyperparameters in the loss function as $\lambda$ = 1, $\mu$ = 0.1, $\gamma$ = 0.01, and $\delta$ = 1.
}

\subsection{Datasets}
\label{sec:datasets}

\jonathan{We used four datasets for our experimental evaluation: PanoContext~\cite{zhang2014panocontext}, Stanford 2D-3D~\cite{armeni2017st2d3d}, and MatterportLayout~\cite{mp3d_dataset} for the panoramic domain, as well as LSUN~\cite{lsun_dataset} for the perspective domain.}
\vspace{-4mm}
\subsubsection*{PanoContext and Stanford 2D-3D}
\jonathan{
The PanoContext~\cite{zhang2014panocontext} and Stanford 2D-3D~\cite{armeni2017st2d3d} datasets contain 514 and 552 annotated cuboid room layouts, respectively, labeled by~\cite{zou2018layoutnet}. We follow the same data splits as LayoutNet~\cite{zou2018layoutnet}, for PanoContext~\cite{zhang2014panocontext}, 413 images for training, 46 for validation, and 53 for testing; and for Stanford 2D-3D~\cite{armeni2017st2d3d}, 404 images for training, 33 for validation, and 113 for testing.
}
\vspace{-4mm}
\subsubsection*{MatterportLayout}
\jonathan{
The MatterportLayout~\cite{mp3d_dataset} dataset, is a subset of the larger Matterport3D dataset as described in~\cite{matterport3d}. It comprises more than 10,800 panorama images. Within this dataset, there are 2,295 real-world room layouts which have been labeled by~\cite{mp3d_dataset}. The data splits of training, validation, and test sets are as follows: 1,647 images are for training, 190 for validation, and 458 for testing.
}
\vspace{-4mm}
\subsubsection*{LSUN}
\jonathan{
The LSUN~\cite{lsun_dataset} datasets, comprising a wide range of FoV and resolution images, consist of 4000 training images, 394 validation images, and 1000 test images. As the ground truth labels for the test set are unavailable, we evaluated our model's performance on the validation set, as has been customary in prior research.
Our model specifically targets the prediction of ceiling and floor boundaries, leading us to exclude images that only contain wall boundaries. As a result, our training set was reduced to 3880 images, and the validation set to 389 images.
}

\jonathan{
As the LSUN~\cite{lsun_dataset} dataset does not provide horizon lines and pitch angles, we validated our vertical shift concept, detailed in~\cref{sec:vertical_shift}, by utilizing the following approach. For images with both ceiling and floor boundaries, we calculated the horizon line by averaging the y-axis values of the lowest ceiling boundary and the highest floor boundary in ground truth boundaries. For images with only one boundary, we used perspective fields from~\cite{jin2023perspective} to determine the horizon line and pitch angle, as a single boundary is insufficient. With these values, we utilized our vertical shift approach to align the panoramic and perspective images. This method was applied to both the training and validation sets to ensure a fair comparison.
}


\subsection{Evaluation Metrics}
\label{sec:evaluation}

\jonathan{
In the field of panoramic imaging, we utilize the established evaluation metrics outlined in~\cite{zou2018layoutnet}. This involves computing the horizon depth, which indicates the distance from the camera to the horizon, for both the ground truth and predicted boundaries. We then project these boundaries onto a bird's-eye view plane using a fixed camera height of 1.6 meters. Subsequently, we evaluate the intersection over the union (IoU) of floor shapes (2D IoU) as well as the entire room volume (3D IoU).
}

\jonathan{
In the context of perspective domains, we compute the 2D IoU for both the ceiling and the floor. This approach is crucial because even minor discrepancies in boundary pixels within the image can lead to significant deviations in the 2D IoU when projected onto a bird's-eye view plane. Due to the limited FoV in perspective, the images often capture only a portion of the room, such as the ceiling or floor, containing one boundary. In those instances, we evaluate the 2D IoU for the ceiling and floor separately.
}

\subsection{Experiment Results}
\label{sec:experiment}

\begin{table}[!t]
\centering
\fontsize{8}{12}\selectfont
  \caption{Experiments with the PanoContext~\cite{zhang2014panocontext} + whole Stanford 2D-3D~\cite{armeni2017st2d3d} and LSUN~\cite{lsun_dataset} datasets.}
  \label{tab_exp_pano_wst2d3d}
  \centering
  \centering
  \begin{tabular}{l cc | cc}
    \toprule
    &
    \multicolumn{2}{c}{Pano.~\cite{zhang2014panocontext} +}
    \\
    &
    \multicolumn{2}{c}{Whole St2D-3D.~\cite{armeni2017st2d3d}} &
    \multicolumn{2}{c}{LSUN~\cite{lsun_dataset}}
    \\
    \midrule
        &&& Ceiling & Floor
        \\
        Method 
        & {2D IoU} 
        & {3D IoU} 
        & {2D IoU} 
        & {2D IoU}
        \\
    \midrule
        LGT-Net~\cite{jiang2022lgt}
        & 87.88 & 85.16 & 23.49 & 42.69
        \\
        DOP-Net~\cite{shen2023dopnet}
        & 88.02 & 85.46 & 33.02 & 41.10
        \\
    \midrule
        LSUN-ROOM~\cite{lsun_room_2018}
        & -	    & -	    & 76.59 & 73.62
        \\
        FUSING~\cite{fusing2023}
        & -	    & -	    & 80.68 & 80.03
        \\
    \midrule
        Ours 
        &\textbf{88.70}
        &\textbf{86.04}	&\textbf{83.12}	&\textbf{80.12}
        \\

    \bottomrule
  \end{tabular}
  \vspace{-5mm}
\end{table}

\jonathan{
In this section, we will present the experimental results of uLayout using three different panoramic datasets: PanoContext~\cite{zhang2014panocontext}, Stanford 2D-3D~\cite{armeni2017st2d3d} and MatterportLayout~\cite{mp3d_dataset}, , which were jointly trained with the LSUN dataset~\cite{lsun_dataset}. We will compare these results with two state-of-the-art panoramic baselines, LGT-Net~\cite{jiang2022lgt} and DOP-Net~\cite{shen2023dopnet}, as well as with two state-of-the-art perspective baselines, LSUN-ROOM~\cite{lsun_room_2018} and FUSING~\cite{fusing2023}.
}

\jonathan{
For the PanoContext~\cite{zhang2014panocontext} and Stanford 2D-3D~\cite{armeni2017st2d3d} datasets, we follow LGT-Net~\cite{jiang2022lgt} and DOP-Net~\cite{shen2023dopnet} to adopt the combined panoramic datasets setting mentioned in~\cite{zou2018layoutnet}. The ``PanoContext + whole Stanford 2D-3D'' 
configuration includes the training split of PanoContext~\cite{zhang2014panocontext} and all split of Stanford 2D-3D~\cite{armeni2017st2d3d} dataset. Similarly, the ``Stanford 2D-3D + whole PanoContext''configuration follows the same definition. 
}

\begin{table}[!t]
\centering
\fontsize{8}{12}\selectfont
  \caption{Experiments with the Stanford 2D-3D~\cite{armeni2017st2d3d} + whole PanoContext~\cite{zhang2014panocontext} and LSUN~\cite{lsun_dataset} datasets.}
  \label{tab_exp_st2d3d_wpano}
  \centering
  \centering
  \begin{tabular}{l cc | cc}
    \toprule
    &
    \multicolumn{2}{c}{St2D-3D.~\cite{armeni2017st2d3d} +}
    \\
    &
    \multicolumn{2}{c}{Whole Pano.~\cite{zhang2014panocontext}} &
    \multicolumn{2}{c}{LSUN~\cite{lsun_dataset}}
    \\
    \midrule
        &&& Ceiling & Floor
        \\
        Method 
        & {2D IoU} 
        & {3D IoU} 
        & {2D IoU} 
        & {2D IoU}
        \\
    \midrule
        LGT-Net~\cite{jiang2022lgt}
        & 88.09 & 86.03 & 16.91 & 30.94
        \\
        DOP-Net~\cite{shen2023dopnet}
        & 87.73 & 85.58 & 19.88 & 30.81
        \\
    \midrule
        LSUN-ROOM~\cite{lsun_room_2018}
        & -	    & -	    & 76.59 & 73.62
        \\
        FUSING~\cite{fusing2023}
        & -	    & -	    & 80.68 & 80.03
        \\
    \midrule
        Ours 
        &\textbf{88.64}
        &\textbf{86.90}	&\textbf{83.30}	&\textbf{80.11}
        \\

    \bottomrule
  \end{tabular}
  \vspace{-5mm}
\end{table}
\jonathan{
In~\cref{tab_exp_pano_wst2d3d} and~\cref{tab_exp_st2d3d_wpano}, When comparing ulayout model with two panoramic baselines, it becomes evident that joint training with the whole LSUN dataset~\cite{lsun_dataset} yields a performance boost of over 0.5\% on both 2D IoU and 3D IoU metrics. Moreover, there's a remarkable increase of over 40\% in 2D IoU for ceiling and floor predictions in perspective. This enhancement underscores the effectiveness of incorporating perspective data during training, enabling the model to glean richer insights and address the limitations observed in the current state-of-the-art models in the panoramic domain regarding perspective prediction. Compared to the two perspective baselines, our model also demonstrates superior performance in ceiling and floor 2D IoU.
This improvement is particularly significant as it highlights the model's capability to effectively integrate and leverage both panoramic and perspective data. Such integration enables the model to perform robustly across diverse scenarios, offering a comprehensive solution that bridges the gap between different data modalities. 
}

\jonathan{
In~\cref{tab_exp_mp3d}, while comparing with two panoramic baselines, uLayout's 2D IoU performance is slightly lower than that of DOP-Net~\cite{shen2023dopnet}, but its 3D IoU performance surpasses both panoramic baselines. Consistent with the findings in~\cref{tab_exp_pano_wst2d3d} and~\cref{tab_exp_st2d3d_wpano}, our model shows over a 40\% performance increase compared to two panoramic baselines in perspective. Regarding the two perspective baselines, uLayout outperforms them in ceiling and floor 2D IoU, particularly excelling in ceiling 2D IoU. These results suggest that joint training on panoramic and perspective data can enhance performance across both domains.}

\begin{table}[!t]
\centering
\fontsize{8}{12}\selectfont
  \caption{Experiments with the MatterportLayout~\cite{mp3d_dataset} and LSUN~\cite{lsun_dataset} datasets.}
  \label{tab_exp_mp3d}
  \centering
  \centering
  \begin{tabular}{l cc | cc}
    \toprule
    &
    \multicolumn{2}{c}{MatterportLayout~\cite{mp3d_dataset}} &
    \multicolumn{2}{c}{LSUN~\cite{lsun_dataset}}
    \\
    \midrule
        &&& Ceiling & Floor
        \\
        Method 
        & {2DIoU} 
        & {3DIoU} 
        & {2DIoU} 
        & {2DIoU}
        \\
    \midrule
        LGT-Net~\cite{jiang2022lgt}
        & 83.52 & 81.11 & 6.86 & 26.06
        \\
        DOP-Net~\cite{shen2023dopnet}
        & \textbf{84.11} & 81.70 & 8.78 & 31.17
        \\
    \midrule
        LSUN-ROOM~\cite{lsun_room_2018}
        & -	    & -	    & 76.59 & 73.62
        \\
        FUSING~\cite{fusing2023}
        & -	    & -	    & 80.68 & 80.03
        \\
    \midrule
        Ours 
        & 84.05	
        &\textbf{81.84}	&\textbf{83.61}	&\textbf{80.25}
        \\

    \bottomrule
  \end{tabular}
\end{table}

\jonathan{
Overall, the uLayout model demonstrates strong generalization capabilities across different datasets. By training on the LSUN dataset and evaluating multiple benchmarks, the model consistently shows enhanced performance in both panoramic and perspective domains. This ability to generalize indicates that the model is not only effective in a specific dataset but also adaptable to various data distributions and environments.
}

\subsection{Ablation Study}
\label{sec:ablation}

\begin{table}[!t]
\centering
\fontsize{8}{12}\selectfont
  \caption{Ablation study with the MatterportLayout~\cite{mp3d_dataset} and LSUN~\cite{lsun_dataset} datasets.}
  \label{tab_exp_ablation}
  \centering
  \centering
  \begin{tabular}{l cc | cc}
    \toprule
    &
    \multicolumn{2}{c}{MatterportLayout~\cite{mp3d_dataset}} &
    \multicolumn{2}{c}{LSUN~\cite{lsun_dataset}}
    \\
    \midrule
        &&& Ceiling & Floor
        \\
        Method 
        & {2D IoU} 
        & {3D IoU} 
        & {2D IoU} 
        & {2D IoU}
        \\
    \midrule
        only panorama
        & 83.08 & 80.83 & 3.96 & 0.57
        \\
        only perspective
        & 31.71 & 29.46 & 74.84 & 75.09
        \\
        w/o Vertical Shift
        & 83.35	& 81.32 & 78.22 & 72.20
        \\
    \midrule
        Ours 
        &\textbf{84.05}	
        &\textbf{81.84}	&\textbf{83.61}	&\textbf{80.25}
        \\

    \bottomrule
  \end{tabular}
  \vspace{-3mm}
\end{table}

\jonathan{
In this section, we conduct an ablation study to evaluate both the performance and computational cost of our uLayout model. Specifically, we will assess the model's performance when trained solely on the MatterportLayout~\cite{mp3d_dataset} dataset, solely on the LSUN~\cite{lsun_dataset} dataset, and jointly on both datasets to demonstrate the benefits of joint training across domains. Additionally, we will present results without applying the vertical shift mentioned in~\cref{sec:vertical_shift} to evaluate the effectiveness of this method. From the perspective of computational cost, we will assess the time, memory usage, and Floating-Point Operations (FLOPs) associated with efficient feature extraction (outlined in~\cref{sec:effi_feature_extract}) and compare these metrics with those of the original feature extractor proposed by~\cite{horizonnet} and also applied in LGT-Net~\cite{jiang2022lgt}.
}

\begin{table}[!t]
\centering
\fontsize{8}{12}\selectfont
  \caption{Ablation study about inference time, GPU Memory and FLOPs on perspective image.}
  \label{tab_exp_ablation_mem_flops}
  \centering
  \begin{tabular}{l ccc}
    \toprule
    &
    \multicolumn{3}{c}{Resnet50~\cite{Resnet}}
    \\
    \midrule
        & {time$\downarrow$}
        & {Memory$\downarrow$} 
        & {FLOPs$\downarrow$} 
        \\
        Method
        & {(ms)}
        & {(MB)} 
        & {(GFLOPs)} 
        \\

    \midrule
        Original Feature Extraction
        &1.88 &75.00 &107.93
        \\
        Efficient Feature Extraction
        &1.88 &\textbf{19.22} &\textbf{26.98}
        \\
    \midrule
        &
        \multicolumn{3}{c}{1D Convolution}
        \\
    \midrule
        & {time$\downarrow$}
        & {Memory$\downarrow$} 
        & {FLOPs$\downarrow$} 
        \\
        Method
        & {(ms)}
        & {(MB)} 
        & {(GFLOPs)} 
        \\
    \midrule
        Original Feature Extraction
        &3.48 &1.25 &63.86
        \\
        Efficient Feature Extraction
        &\textbf{1.17} &\textbf{0.31}
        &\textbf{15.97}
        \\
    \bottomrule
  \end{tabular}
 \vspace{-3mm}
\end{table}
\jonathan{
In~\cref{tab_exp_ablation}, training solely on panorama images results in high performance for panorama views but relatively low performance for perspective views. Conversely, when trained exclusively on perspective images, the model demonstrates poor performance on panorama views but performs reasonably well on perspective views. These findings suggest that specialized training on a single type of imagery enhances performance for that type but does not generalize well across different image domains. Excluding the vertical shift method results in a noticeable drop in performance across both domains due to an ill-posed issue in perspective, as described in~\cref{sec:vertical_shift}. Without proper vertical alignment, the model struggles to seamlessly learn features from both domains, leading to reduced overall effectiveness.
In contrast, our proposed model, which incorporates both panorama and perspective training along with the vertical shift, achieves the best results. This demonstrates the effectiveness of our approach in leveraging multiple views and the vertical shift technique for improved layout estimation.
}

\jonathan{
In the ``w/o Vertical Shift'' experiment, we exclude the depth loss, normal loss, and gradient loss from the loss setting mentioned in~\cref{sec:prediction_and_loss}. The rationale is that projecting the image domain boundaries to the horizon depth in a bird's-eye view plane could result in infinite or even negative horizon depth values. Additionally, for the same reason, we still apply the Vertical Shift on the validation set in LSUN~\cite{lsun_dataset} when calculating the ceiling and floor 2DIoU.
}

\jonathan{
In~\cref{tab_exp_ablation_mem_flops}, we present measurements of time, memory, and FLOPs in the perspective domain for both the original feature extractor proposed in~\cite{horizonnet} and the efficient feature extractor in our uLayout model. Our design adjustments in~\cref{sec:effi_feature_extract} result in a remarkable reduction of approximately 75\% in both memory usage and FLOPs for Resnet-50~\cite{Resnet}. This reduction is achieved by minimizing the zero-padding region by 75\% on perspective data. Similarly, in the convolution part, we've managed to reduce time, memory, and FLOPs by nearly 75\%.
}
\begin{figure*}[t!]
\centering
  \includegraphics[ width=\linewidth]
    {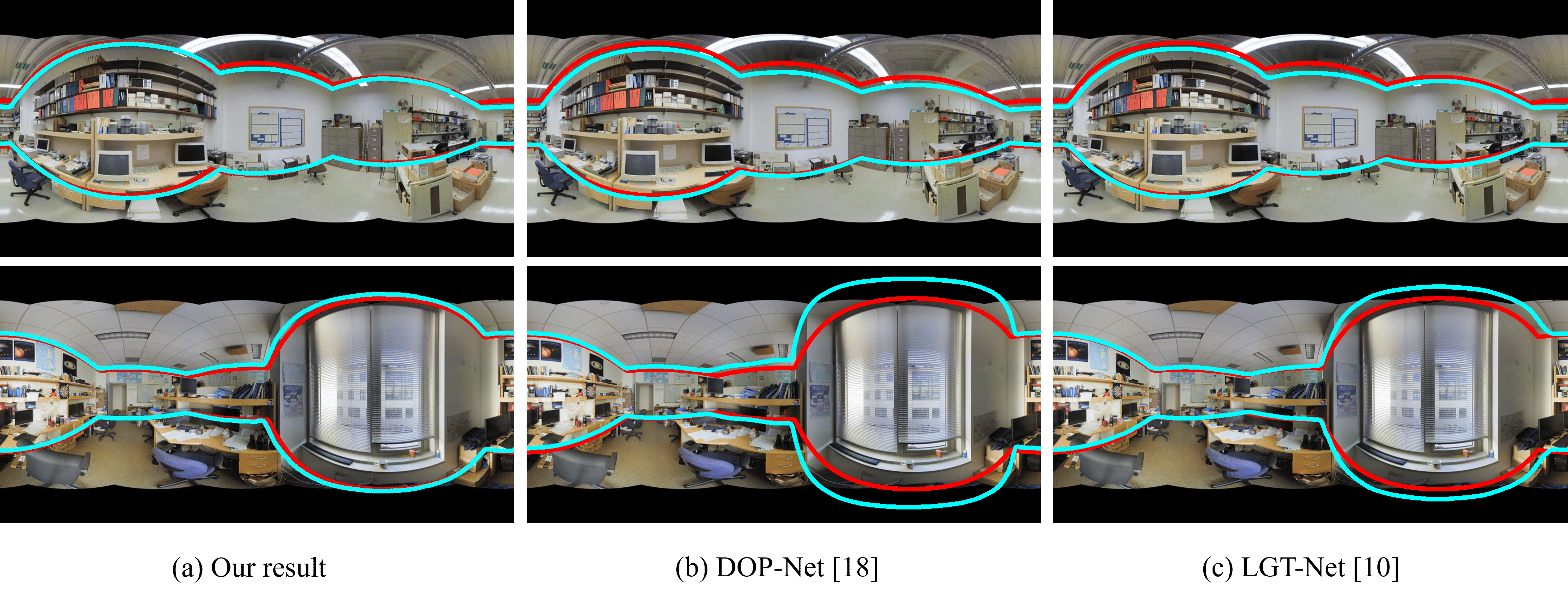}
    \vspace{-7mm}
    \caption{\textbf{Qualitative Results} for Panoramic Images. Red line denote ground truth layout. Cyan lines denote predicted layout.}
    \label{fig_pano_qualitative}    
\end{figure*}
\begin{figure*}[t!]
\centering
  \includegraphics[ width=\linewidth]
    {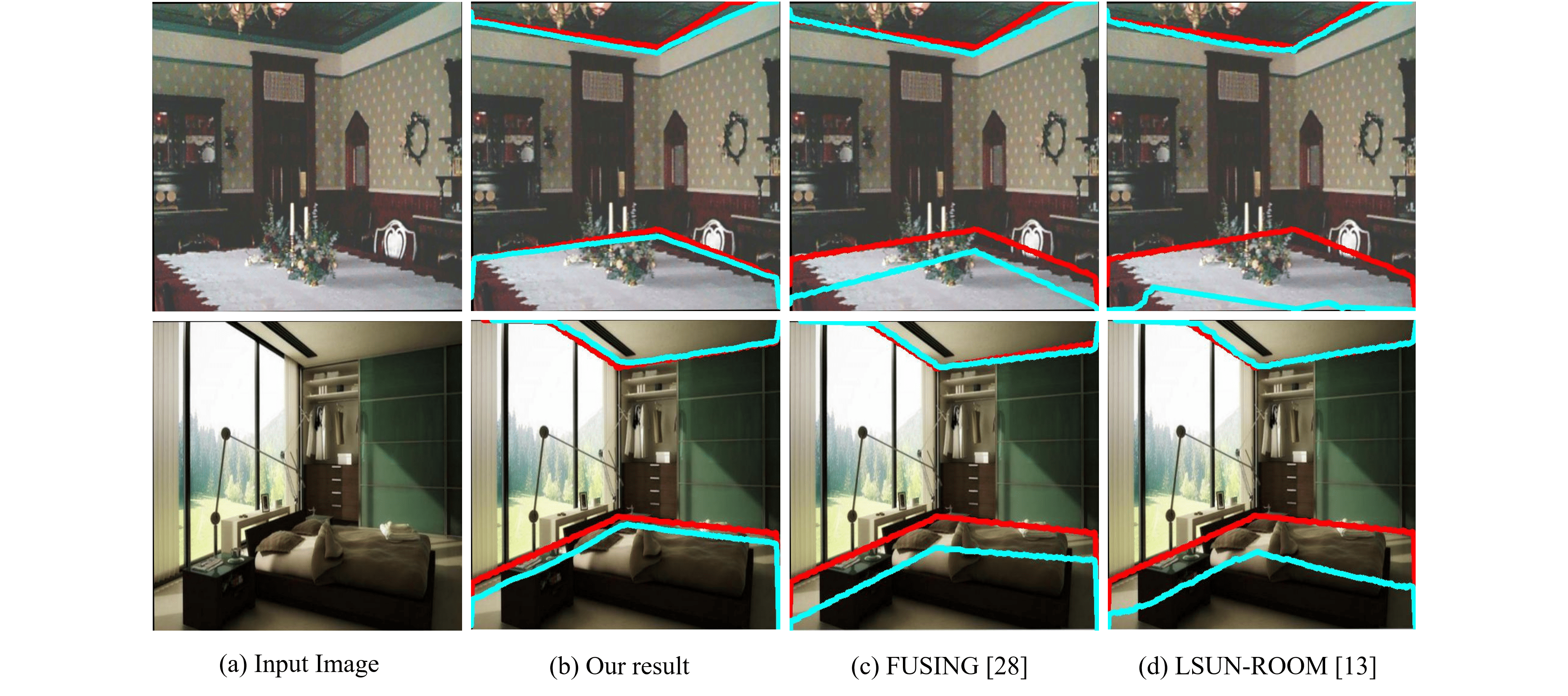}
    \vspace{-7mm}
    \caption{\textbf{Qualitative Results} for Perspective Images. Red line denote ground truth layout. Cyan lines denote predicted layout.}
    \label{fig_pp_qualitative}
    \vspace{-4mm}    
\end{figure*}
\subsection{Qualitative Results}
\label{sec:qualitative}

\subsubsection{Qualitative Result on Panoramic Images.}

\jonathan{
In \cref{fig_pano_qualitative}, we present several qualitative samples of panoramic images from the Stanford 2D-3D~\cite{armeni2017st2d3d} dataset. Our model shows effective improvements compared to the two panoramic baselines, LGT-Net~\cite{jiang2022lgt} and DOP-Net~\cite{shen2023dopnet}, by predicting more precise boundaries when the camera is close to the wall.
}
\vspace{-2mm}
\subsubsection{Qualitative Result on Perspective Images.}

\jonathan{
We present several qualitative results from our uLayout model alongside two perspective baselines. Our model demonstrates effective improvement in predicting scenes where boundaries are occluded by furniture in \cref{fig_pp_qualitative}.
}

\section{Conclusion}
\jonathan{
In this paper, we employ a unified model capable of joint training on panorama and perspective images. Additionally, we propose a vertical shift approach to transform perspective images into a format more similar to panoramas, facilitating easier learning for the model across both domains. Moreover, we introduce efficient feature extraction techniques to reduce computation time, memory usage, and FLOPs in the perspective domain. Through comprehensive evaluation, we present a state-of-the-art unified model proficient in handling data from both perspective and panorama domains, demonstrating promising performance.
}
\section*{Acknowledgments}
\jonathan{
This project is supported by The National Science and Technology Council NSTC and The Taiwan Computing Cloud TWCC under the project NSTC 113-2221-E-007-105-MY3.
}

{\small
\bibliographystyle{ieee_fullname}
\bibliography{main}
}

\end{document}